# Improving code-mixed hate detection by native sample mixing: A case study for Hindi-English code-mixed scenario

DEBAJYOTI MAZUMDER, Indian Institute of Science Education and Research Bhopal, India
AAKASH KUMAR, Indian Institute of Science Education and Research Bhopal, India
JASABANTA PATRO, Indian Institute of Science Education and Research Bhopal, India

Hate detection has long been a challenging task for the NLP community. The task becomes complex in a code-mixed environment because the models must understand the context and the hate expressed through language alteration. Compared to the monolingual setup, we see much less work on code-mixed hate as large-scale annotated hate corpora are unavailable for the study. To overcome this bottleneck, we propose using native language hate samples (native language samples/ native samples hereafter). We hypothesise that in the era of multilingual language models (MLMs), hate in code-mixed settings can be detected by majorly relying on the native language samples. Even though the NLP literature reports the effectiveness of MLMs on hate detection in many cross-lingual settings, their extensive evaluation in a code-mixed scenario is yet to be done. This paper attempts to fill this gap through rigorous empirical experiments. We considered the Hindi-English code-mixed setup as a case study as we have the linguistic expertise for the same. Some of the interesting observations we got are: (i) adding native hate samples in the code-mixed training set, even in small quantity, improved the performance of MLMs for code-mixed hate detection, (ii) MLMs trained with native samples alone observed to be detecting code-mixed hate to a large extent, (iii) the visualisation of attention scores revealed that, when native samples were included in training, MLMs could better focus on the hate emitting words in the code-mixed context, and (iv) finally, when hate is subjective or sarcastic, naively mixing native samples doesn't help much to detect code-mixed hate. We will release the data and code repository to reproduce the reported results.

## 1 INTRODUCTION:

The rising cases of online hate-speech [27] and similar components (cyberbullying [18], racism [57], gender discrimination [29], radicalisation [56], religious hatred, abusive language detection [3], etc.) distress the sanity and the civic nature of online discussions. To address this, the NLP community have been working long on characterisation, detection and mitigation of these components [1, 9, 20, 25, 27, 44, 45, 51, 54]. The intensity of their focus can be gauged by the fact that around 460+ peer-reviewed AI/ML papers were published between 2001 and 2021 [27] on this topic,

Authors' addresses: Debajyoti Mazumder, debajyoti22@iiserb.ac.in, Indian Institute of Science Education and Research Bhopal, Bhopal Bypass Road, Bhauri, Bhopal, Madhya Pradesh, India, 462066; Aakash Kumar, aakashk19@iiserb.ac.in, Indian Institute of Science Education and Research Bhopal, Bhopal Bypass Road, Bhauri, Bhopal, Madhya Pradesh, India, 462066; Jasabanta Patro, jpatro@iiserb.ac.in, Indian Institute of Science Education and Research Bhopal, Bhopal Bypass Road, Bhauri, Bhopal, Madhya Pradesh, India, 462066.

and exponentially rising since then. The community has proposed 70+ datasets spanning 20+ languages and different modalities (memes, texts, social media posts, images, videos, etc.) [9, 27].

Till now, we have yet to eradicate the hate completely from the online platforms. The main reason is that most of the studies were done for a few resource-rich languages (51% alone for English), whereas we have 100+ languages in the world, each having seven million+ speakers[1]. Further, a small fraction of these studies focused on code-mixed setups where the hate is uttered altering more than one language. It is common in multilingual environments like Europe, India and Latin America, where a significant portion of the population knows more than one language. Hate detection in code-mixed language is more complex compared to the monolingual environment. The models must understand the context and the hate expressed through multiple languages. One of the primary bottlenecks for the research on code-mixed hate detection is the unavailability of large-scale training corpora. The same reason also holds for most of the low-resource languages.

The emergence of multilingual language models (MLMs) allowed us to address this issue via cross-lingual learning. It enables the models to learn task-specific knowledge from a dataset in one language and make predictions for samples in different languages. Past works demonstrated that MLMs, to a significant extent, could detect hate when train and test languages are different [26]. This is because (i) the vocabulary of MLMs covers many languages, and (ii) their embeddings encode the semantic and syntactic features seen across multiple languages. However, they face several criticisms too. The NLP literature reported [6, 26] that cross-lingual learning fails when hate (i) expresses language-specific taboos and (ii) is specific to a community or culture. We argue that these limitations will not likely hold in the code-mixed setup if we rely on native samples. Our intuition is that the native corpora will capture both. This paper attempts to showcase it by doing empirical experiments. Particularly, we attempted to answer the following two research questions,

- **R1:** *Does training with additional native hate samples impact the code-mixed hate detection?*
- **R2:** *Can training with only native samples detect the hate in a code-mix scenario?*

We have done all of our experiments for the Hindi-English code-mixed environment. This is because (i) we have the necessary linguistic expertise for this pair ( It is required for calculating code-mixing complexity (section 5), and error analysis (section 7)), and (ii) we could find annotated code-mixed and respective native language hate corpora publicly available (GPL 3). Our contributions in this paper can be summarised as,

- We evaluated the impact of native language hate samples on code-mixed hate detection (**Exp 1:** Section 4.3) by comparing the performance across three training sets, (i) having only code-mixed hate samples, (ii) with additional samples from Hindi and English hate corpora, and (iii) with only Hindi and English samples. We experimented with three types of models: (i) statistical classifiers on top of the word n-grams, (ii) fine-tuning MLMs such as mBERT [17], XLM [35], XLM-R [12], IndicBERT [30] and MuRIL [32], and (iii) Fine-tuning MLMs with additional transformer layers. Further, we reported the model performance when native samples were added, with an equal label ratio or with a ratio the same as in the code-mixed hate corpora. Furthermore, we also reported the performance variations of MLMs when native samples were added in different amounts (**Exp 2:** section 4.4).
- We evaluated the performance of MLMs by training them only with native samples (**Exp 3:** Section 4.5) as well. We created three types of training sets: (i) with only Hindi samples, (ii) with only English samples and (ii) with Hindi and English samples together.

[1] http://www2.harpercollege.edu/mhealy/g101ilec/intro/clt/cltclt/top100.html

- We visualised the attention scores given by the MLMs and reported the change in scores on hate-emitting words after native samples were mixed with the original code-mixed training set (see Figure 2 in appendix).
- Finally, we manually inspected each test prediction and reported the cases for which MLMs (a) got better, (b) remained confused, and (c) performed worse after native samples were mixed in the code-mixed training set.

Some of the interesting results we got are,

- On combining the native samples, while the statistical models performed worse, many MLMs reported significant ($p < 0.05$) improvements (an increase of **0.11** in $F1$ score) in the best performing model.
- MLMs, when trained with only native samples, could identify hate in a code-mixed context nearly as good ($F1$ score: **0.60** ) as when trained with additional code-mixed samples ( $F1$ score: **0.63**). It implies that we can deploy the MLMs trained with native samples if an appropriate code-mixed corpus is unavailable. Further, we also observed that the native language dataset that shares maximum linguistic overlap with the code-mixed hate corpus is more prominent in capturing the code-mixed hate.
- We observed that after native sample mixing, the MLMs gave high attention scores to the hate-emitting words when they appeared in the code-mixed context. Some of such cases were demonstrated in Figure 2.
- Finally, in the error analysis, we also reported cases where hate is sarcastic and subjective; it's still hard to identify them in the code-mixed setting by simply relying on native hate samples.

## 2 RELATED WORKS:

### 2.1 Monolingual hate:

In the monolingual set-up, detecting and mitigating hate and similar components were extensively studied in the literature [9, 25, 27]. Still, as mentioned earlier, the language coverage is yet to rise significantly. From a task formulation point of view, we saw several variations, such as binary, multi-class, and multi-label classification. There were several datasets proposed that did binary categorization of samples such as: 'hate' or not, [4, 21, 46], 'harassment' or not [23], 'personal attack' or not [59], 'sexism' or not [29], 'offensive' or not [2], 'anti-refugee hate' or not [49, 50] and 'islamophobic' or not [11]. Similarly, from a multi-class and multi-label perspective, researchers combined previously mentioned labels; datasets were annotated with: 'sexist'/'racist'/'none' [58], 'obscene'/'offensive but not obscene'/'clean' [39], hate against 'gender'/'sexual orientation'/'religion/'disability'/'none' [24, 42, 47] etc. From an architecture point of view, over the years, we observed a general shift from traditional rule-based and keyword-based systems to machine-learning methods (SVM, Random forest, etc.) and then to deep-neural methods(linear layers, RNNs, LSTMs, CNNs, Transformers)[9, 25, 27]. With the rise of transformer-based language models, fine-tuning their last layers or using neural layers on top of them got approved as a standard go-to approach. Meanwhile, in parallel, researchers also experimented with strategies like ensemble methods, multi-tasking and transfer learning [9, 25, 27] frameworks. These methods produced state-of-the-art results in their respective timelines. For model explainability, people focused on projecting attention scores on input segments to check the model's focus on hate words and hate targets [37].

### 2.2 Code-mixed hate:

Very few studies focused on hate detection in code-mixed environments [7, 10, 16, 31, 34, 38]. Most of them focused on the Hindi-English code-mixed scenario. From a task formulation point of view, we also saw several variations. In binary classification framework, Bohra et al. [7] presented a dataset of 4.8k tweets, each annotated with 'hate'/'non-hate'. In multi-class and multi-label set-ups, Kumar et al. [34] published a dataset of 21k Facebook posts, each annotated with one of three labels ('covert aggression'/ 'overt aggression'/ 'none') and their associated category ('physical threat', 'sexual threat', 'identity threat', 'non-threatening aggression'). Similarly, Mathur et al. [38] produced a dataset focusing only on sexism tweets, each annotated with 'not-offensive'/ 'abusive'/ 'hate'. Approach-wise, we see people have relied upon (i) feature-based methods (n-grams, lexicons, negations, etc.) with traditional machine learning algorithms like SVM, Random forest, etc. [7], and (ii) neural architectures like CNNs, and LSTMs on top of LIWC features and sentiment scores [38].

### 2.3 Hate detection by cross-lingual learning:

In the context of using cross-lingual learning for hate detection, Bigoulaeva et al. [6] demonstrated that neural architectures like CNNs and LSTMs, without using pre-trained MLMs, could detect hate expressed in German text after training was done with the English hate corpora. However, they additionally used an unlabelled German corpora. Firmino et al. [19], on the other hand, showed that training MLMs like mBERT and XLM with English and Italian hate corpora can detect the hate in Portuguese samples. However, as mentioned earlier, previous work reported that cross-lingual learning fails when hate (i) expresses language-specific taboos and (ii) is specific to a community or culture.

### 2.4 Research gap:

We couldn't find any critical study evaluating impact of native hate samples on code-mixed hate detection. We argue that limitations of cross-lingual learning[41] for hate detection, like failing to identify language-specific taboos and culture-specific hate, may not appear in the code-mixed setup. Our intuition is that the native corpora will be capturing both. As an initial attempt to show the impact of native samples on code-mixed hate detection, we considered a simple binary hate classification framework (i.e. 'hate' or not), in a Hindi-English code-mixed setting.

## 3 DATASET DETAILS:

We considered the Hindi-English code-mixed hate dataset [8], made up of social media posts. The authors retrieved 1,12,718 tweets based on a list of hashtags and keywords related to politics, public protests, riots, etc. Out of which, they manually filtered 4575 code-mixed tweets. Two linguistic experts then manually annotated the filtered tweets as "hate" or "non-hate". Additionally, for native language samples, we considered Hindi [15] and English [2] hate datasets in our experiments. We chose them as they are (i) relatively balanced, (ii) versatile, and (iii) widely used in past works [14, 40, 44]. Their samples are manually annotated as "hate"/ "non-hate". The statistical distribution of class labels and the Kullback-Leibler (KL) divergence [33] for the datasets are reported in Table 1.

[2]HASOC-2019: https://hasocfire.github.io/hasoc/2019/dataset.html

**Table 1.** Distribution of labels and KL divergence present in the considered code-mixed and monolingual hate corpora.

| Dataset | Hate | Non-Hate | KL |
|---|---|---|---|
| English-Hindi CM[8] | 1661 | 2914 | 1.218 |
| English[2] | 2261 | 3591 | 1.182 |
| Hindi[15] | 3338 | 1416 | 0.740 |

The KL divergence measures the lexical variation in samples belonging to two classes. Statistical classifiers like Naive Bayes are known to perform better in datasets where KL divergence is high [5]. Table-2 presents some samples from all three considered datasets. The datasets are open-source i.e. two of them are GPL-3 licensed, the other can be easily collected by simple sign-up.

**Table 2.** Samples from all three considered datasets. The Hindi parts (Romanized or Devanagri) are underlined for identification. We have provided English translation of the code-mixed and Hindi samples for readability.

| Classes | English-Hindi code-mixed |
|---|---|
| Hate | Aise logo se sakht nafrat karta hu Jo caste ko naam ke sath jod ke chaude hote h but real me vo piddu hote h |
| | **Translation:** I strongly hate those people who associate caste with their names and act tough, but in reality, they are weak. |
| Non-Hate | Sir phansi nahi. .. sirf looted money wapas chaiya |
| | **Translation:** Sir, not execution, we just want the looted money back. |

| Classes | English |
|---|---|
| Hate | I don't know how much more I can take! 45 is a compulsive liar! #Trump30Hours #TrumpIsATraitor |
| Non-Hate | Stand by the doctors , with the doctors. They need support and world needs to interfere. |

| Classes | Hindi |
|---|---|
| Hate | दलित बिलकुल नीच होते हैं |
| | **Translation:** Dalits are downright mean. |
| Non-Hate | हमें महिलाओं का प्रोत्साहन करना चाहिए |
| | **Translation:** We should encourage women. |

## 4 EXPERIMENTS:

### 4.1 Experimental set-up:

We considered three generic architectures for our experiments. Note that our aim was not to propose any novel architecture to form a new state-of-art. Rather, we evaluated the behaviour of widely used statistical and MLM-based models when native samples were added in the training process. A graphical presentation of this concept is presented in figure 1(a). The three generic architectures we considered are,

(1) Statistical classifiers SVM, Random Forest and Naive Bayes on top of the word n-gram features. We considered word-level unigrams, bigrams and trigrams as features, as past work [8] reported them to perform best. Figure 1(b) presents a graphical representation of the same.

(2) As a second approach, we fine-tuned five MLMs mBERT [17], XLM-R [12], XLM [35], IndicBERT [30] and MuRIL [32]. While mBERT and XLM-R were trained with corpora covering 100+ languages and XLM with corpora covering 15 languages, IndicBERT and MuRIL were exclusively trained with Indian languages. We believe comparing both types of MLMs will illustrate the efficacy of MLMs in code-mixed hate detection when they are trained with generic (100+ languages in mBERT and XLM-R and 15 languages in XLM) vs regional (Indian languages) language corpora. The considered pre-trained models were widely used in past works [26, 43, 61]. The samples were passed through respective MLM tokenizers before being utilised for training. Also,

**Fig. 1.** Graphical representations of the training process and the model architectures. Sub-figure (a) presents the overall thought behind the training process. Sub-figure (b), (c) and (d) presents the architectures behind statical models, MLMs and MLMs with transformer layers. In statistical models, the input unigrams, bigrams and trigrams are represented by layers n=1, n=2 and n=3, respectively. $w_i$ represents ith word and $w_i : w_j$ represents an n-gram spanning $w_i$ to $w_j$. The MLM in sub-figure (c) and (d) can be any one of mBERT [17], XLM-R [12], XLM [35], IndicBERT [30] and MuRIL [32].

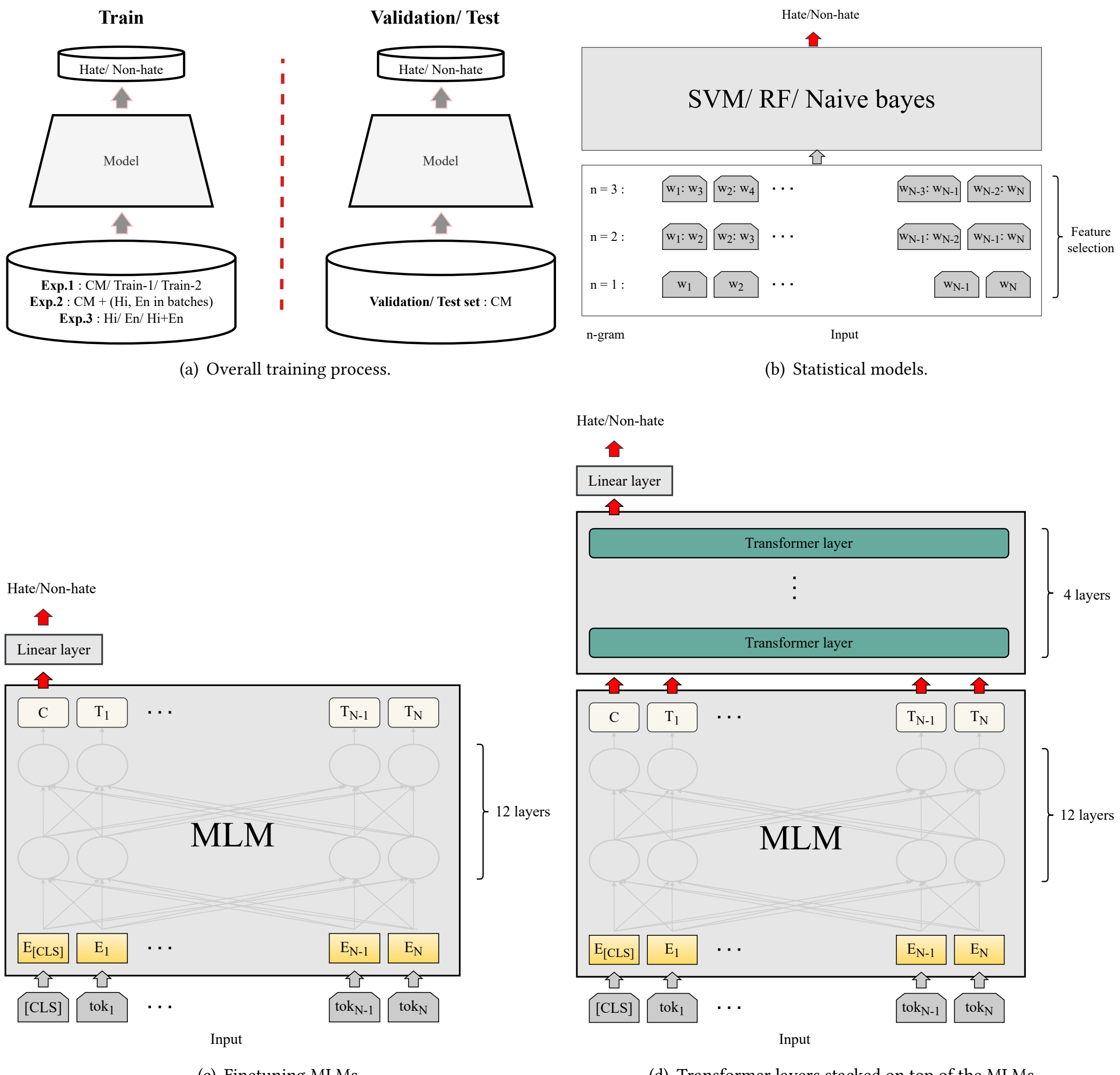

(a) Overall training process.

(b) Statistical models.

(c) Finetuning MLMs.

(d) Transformer layers stacked on top of the MLMs.

We stacked a linear layer on top of the sentence representation embedding ([CLS]) and fine-tuned the last two layers of the MLMs. Figure 1(c) presents a graphical representation of the same.

(3) Lastly, we stacked four transformer layers followed by a linear layer on top of the considered MLMs (represented by mBERT$_{trans}$, XLM-R$_{trans}$, XLM$_{trans}$, IndicBERT$_{trans}$ and MuRIL$_{trans}$ respectively). As previously mentioned, samples were tokenized by the respective MLM tokenizers before being used for training. During fine-tuning, we froze all but the last two layers of the language model. Figure 1(d) presents a graphical representation of this process. Past studies [28, 48] reported that transformer-based language models learn and represent lexical, syntax, and semantic features in their lower, middle, and higher layers. Adding more transformer layers can create more parameter space where MLMs can store task-specific complex semantic information as latent features [52, 53, 60] during training. Thus, experimenting with an additional transformer layer illustrates if storing such complex information is helpful in code-mixed hate detection.

### 4.2 Experimental setting:

For the experiments, we collected the pre-trained weights for the MLMs from the Huggingface[3] transformer library. We experimented with different optimizers, learning rates, schedulers, batch sizes and early stopping strategies. The list of hyper-parameters and their values are listed in Table 3. We found that AdamW optimizer [36], with a learning rate of $2 \times 10^{-5}$, and a scheduler with a learning rate decay of 0.9 (gamma value), performed best. Models were trained for a maximum of 25 epochs with an epoch patience of four. We reported the average F1 score across three random seeds.

**Table 3.** List of different hyper-parameters and their values we considered in our experiments.

| **Parameters** | **Values** |
|---|---|
| Learning rate | {2e-6, 2e-5, 2e-4, 3e-3, 9e-3, 1e-2} |
| Optimizer | SGD, Adam, AdamW |
| Gamma value (scheduler) | 0.9, 0.8, 0.6, 0.4 |
| Batch size | 16, 32, 64 |
| Sequence length | 64, 128, 248 |
| Patience (Early stop) | 2, 4, 6 |

### 4.3 Exp. 1- Impact of added native samples:

In the first experiment, we measured the impact of adding native samples on code-mixed hate detection. First, we created training (70%), validation (15%) and test (15%) splits of the code-mixed hate dataset [8] using stratified sampling. Additionally, we formulated two new training sets by including samples from the English and Hindi hate corpora. In the first set (**Train-1**), we added an equal proportion of “hate” or “non-hate” samples from the monolingual datasets. In the second (**Train-2**), we kept label distribution the same as we observed in the code-mixed training split. The KL divergences [33] and the label distributions present in the formulated training sets, in the initial code-mixed training set, and in the validation and test sets are reported in table 4.

### 4.4 Exp. 2- Impact of native samples added in different amounts:

Our second experiment quantified the impact of English and Hindi samples when they were added in different amounts to the code-mixed training set. For this purpose, we created two sets of training sets. The first set of training sets was created by incrementally adding two batches of native samples, one in English and the other in Hindi, to the code-mixed training set. Each batch had 200 randomly selected samples with an equal ratio of “hate” and “non-hate” samples (similar to **Train-1** in Exp-1). The other set of training sets was similar to the first set, except here, the label ratio in each batch was the same as in the original code-mixed training set (similar to **Train-2** in Exp-1). The distributions of labels in the formulated training sets were reported in Table 5.

[3]https://huggingface.co/

**Table 4.** The distribution of labels present in the formulated training sets **Train-1** (equal ratio) and **Train-2** (code-mixed ratio). The label distribution in the initial code-mixed training set reported next to **CM** in Train-1 and Train-2. The numbers next to **Validation** and **Test** represent the distribution of labels in code-mixed validation and test splits, respectively.

| Partition | Dataset | Hate | Non-hate | KL |
|---|---|---|---|---|
| Train-1 | CM | 1149 | 2062 | 1.333 |
| | Hindi | 1416 | 1416 | 1.413 |
| | English | 1416 | 1416 | 1.539 |
| | Total | 3981 | 4894 | 1.220 |
| Train-2 | CM | 1149 | 2062 | 1.333 |
| | Hindi | 810 | 1416 | 1.833 |
| | English | 2000 | 3500 | 1.210 |
| | Total | 3959 | 6978 | 1.231 |
| Validation | CM | 249 | 438 | 1.977 |
| Test | CM | 249 | 437 | 1.978 |

### 4.5 Exp. 3- Relying only on native samples:

Our last experiment evaluated the performance of MLMs when they were trained with only native samples. The intuition behind this experiment was to check if knowledge learned from the training of native samples was enough to detect hate in the code-mixed scenario. Again, we created three training sets: (i) with only Hindi samples, (ii) with only English samples, and (iii) with English and Hindi samples together. All the training sets had an equal proportion of "hate" and "non-hate" samples. We fine-tuned the hyper-parameters on the code-mixed validation set. In table 6, we reported the label distribution in newly created training sets.

## 5 CODE-MIXING COMPLEXITY:

To understand the code-mixing complexity associated with the Hindi-English hate dataset, we calculated the average code-mixing index (CMI) [13] and the average burstiness [22] index values for the test set. These two metrics generally measure the linguistic complexity of code-mixed text. The CMI for an input text is calculated as,

$$CMI = \begin{cases} 100 * \left[1 - \frac{max(t_i)}{N-I}\right] & \text{for } N > I \\ 0 & \text{for } N = I \end{cases}$$

Where $t_i$ denotes the word count for language $i$, $N$ represents the total word count, and $I$ signifies the number of language-independent words. The CMI index (range: [0,50]) reflects the degree of code-mixing present in the text. A higher CMI index value indicates a greater amount of code-mixing. Specifically, it denotes a fraction of the total words that belong to languages other than the most dominant language in the text.

Similarly, the burstiness index is measured by,

$$Burstiness = \frac{\sigma_Y - m_Y}{\sigma_Y + m_Y}$$

**Table 5.** The distribution of labels present in the newly formulated training sets as a part of Exp-2. The two sets of training sets are represented by column names "Equal ratio" and "CM ratio". $Train_x$ refers to the training set created by adding x native samples to the code-mixed training set.

| Partition | Dataset | Equal ratio | | CM ratio | |
|---|---|---|---|---|---|
| | | Hate | Non-hate | Hate | Non-hate |
| $Train_{200}$ | CM | 1149 | 2062 | 1149 | 2062 |
| | Hindi | 100 | 100 | 73 | 127 |
| | English | 100 | 100 | 73 | 127 |
| | Total | 1349 | 2262 | 1349 | 2262 |
| $Train_{400}$ | CM | 1149 | 2062 | 1149 | 2062 |
| | Hindi | 200 | 200 | 146 | 254 |
| | English | 200 | 200 | 146 | 254 |
| | Total | 1549 | 2462 | 1549 | 2462 |
| $Train_{600}$ | CM | 1149 | 2062 | 1149 | 2062 |
| | Hindi | 300 | 300 | 218 | 382 |
| | English | 300 | 300 | 218 | 382 |
| | Total | 1749 | 2662 | 1749 | 2662 |
| $Train_{800}$ | CM | 1149 | 2062 | 1149 | 2062 |
| | Hindi | 400 | 400 | 291 | 509 |
| | English | 400 | 400 | 291 | 509 |
| | Total | 1949 | 2862 | 1949 | 2862 |
| $Train_{1000}$ | CM | 1149 | 2062 | 1149 | 2062 |
| | Hindi | 500 | 500 | 364 | 636 |
| | English | 500 | 500 | 364 | 636 |
| | Total | 2149 | 3062 | 2149 | 3062 |
| $Train_{1200}$ | CM | 1149 | 2062 | 1149 | 2062 |
| | Hindi | 600 | 600 | 436 | 764 |
| | English | 600 | 600 | 436 | 764 |
| | Total | 2349 | 3262 | 2349 | 3262 |
| $Train_{1400}$ | CM | 1149 | 2062 | 1149 | 2062 |
| | Hindi | 700 | 700 | 509 | 891 |
| | English | 700 | 700 | 509 | 891 |
| | Total | 3549 | 3462 | 3549 | 3462 |
| Validation | CM | 249 | 438 | 249 | 438 |
| Test | CM | 249 | 437 | 249 | 437 |

**Table 6.** The distribution of labels present in the training sets created as a part of experiment-3.

| Partition | Dataset | Hate | Non-hate |
|---|---|---|---|
| Train-1 | English | 1416 | 1416 |
| Train-2 | Hindi | 1416 | 1416 |
| Train-3 | English | 1416 | 1416 |
| | Hindi | 1416 | 1416 |
| | Total | 2832 | 2832 |
| Validation | CM | 249 | 438 |
| Test | CM | 249 | 437 |

Where $\sigma_\gamma$ denotes the standard deviation of language spans and $m_\gamma$ denotes the mean of language spans. The burstiness measure serves as an indicator of code-switching intensity, with a low score (approaching -1) suggesting uniform code-switching patterns and a high score (approaching 1) indicating bursty switching patterns. Both of the measures rely on word-level language tags. As the tags were not readily available, we manually annotated them by two students who are well-versed in Hindi and English. The results for both measures are presented in table 7. They indicate that, on average, samples have significantly high, i.e. 39% words from the non-dominant language (here English) with uniform code-switching patterns.

**Table 7.** Code-mixing complexity measures for test set. **CMI** and **Burst.** refers to the average code mixing index score and burstiness index score for the test set respectively.

| **No. of Samples** | **CMI** | **Burst.** |
|---|---|---|
| Test: 686 | 39 | -0.32 |

## 6 RESULTS AND DISCUSSION:

### 6.1 Exp.-1 observations:

We reported the results of our Exp-1, i.e. measuring the impact of native hate samples on code-mixed hate detection in Table 8. We compared the performance of different models in terms of accuracy (*Acc*), precision (*Pre*), recall (*Rec*) and F1-score ($F1$). The best scores for individual training scenarios were marked in bold. Since the label distributions in all training sets are unbalanced, we believe the $F1$ score best measures the performance. Following were our takeaways,

- When training was done with only code-mixed samples, among statistical models, Naive Bayes reported the highest $F1$-score (**0.51**) followed by SVM (**0.45**). Random Forest performed least with an $F1$-score of **0.23**.

**Table 8.** The results of our experiment measuring the impact of native hate samples on code-mixed hate detection. The evaluating parameters are accuracy(*Acc*), precision(*Pre*), recall(*Rec*) and F1-score($F1$). The scores under columns **Code-mixed**, **Train-1**, and **Train-2** reported the performance when training was done with the original code-mixed training set, and additionally formulated training sets explained in section 4.3. The variations in $F1$ scores for different random seeds were mentioned in brackets.

| **Models** | **Codemix** | | | | **Train-1** | | | | **Train-2** | | | |
|---|---|---|---|---|---|---|---|---|---|---|---|---|
| | **Acc** | **Pre** | **Rec** | **F1** | **Acc** | **Pre** | **Rec** | **F1** | **Acc** | **Pre** | **Rec** | **F1** |
| SVM | 0.67 | 0.58 | 0.37 | 0.45 (±0.03) | 0.65 | 0.51 | 0.39 | 0.44 (±0.01) | 0.69 | 0.60 | 0.40 | 0.48 (±0.01) |
| RF | 0.67 | **0.77** | 0.13 | 0.23 (±0.01) | 0.67 | **0.71** | 0.16 | 0.26 (±0.00) | 0.66 | **0.79** | 0.10 | 0.18 (±0.02) |
| Naive Bayes | 0.63 | 0.49 | 0.53 | 0.51 (±0.02) | 0.66 | 0.65 | 0.16 | 0.26 (±0.02) | 0.67 | 0.62 | 0.25 | 0.36 (±0.01) |
| XLM | 0.68 | 0.52 | **0.56** | 0.52 (±0.01) | 0.69 | 0.58 | 0.47 | 0.52 (±0.01) | 0.67 | 0.53 | 0.63 | 0.58$^*$ (±0.01) |
| mBERT | 0.61 | 0.44 | 0.55 | 0.49 (±0.01) | **0.71** | 0.62 | 0.51 | 0.56$^*$ (±0.02) | 0.68 | 0.57 | 0.55 | 0.56 (±0.00) |
| XLM-R | 0.67 | 0.51 | 0.43 | 0.47 (±0.02) | 0.67 | 0.54 | 0.61 | 0.58$^*$ (±0.01) | 0.69 | 0.60 | 0.46 | 0.52 (±0.01) |
| IndicBERT | 0.65 | 0.48 | 0.47 | 0.48 (±0.05) | 0.67 | 0.55 | 0.56 | 0.55$^*$ (±0.02) | 0.72 | 0.68 | 0.43 | 0.53 (±0.02) |
| MuRIL | 0.68 | 0.52 | **0.56** | **0.54** (±0.03) | 0.64 | 0.50 | 0.70 | 0.58 (±0.01) | 0.66 | 0.53 | **0.69** | 0.60$^*$ (±0.02) |
| XLM$_{trans}$ | 0.65 | 0.48 | 0.52 | 0.50 (±0.02) | 0.69 | 0.56 | 0.63 | 0.59$^*$ (±0.01) | 0.67 | 0.53 | 0.65 | 0.58$^*$ (±0.01) |
| mBERT$_{trans}$ | 0.72 | 0.62 | 0.39 | 0.48 (±0.01) | 0.67 | 0.54 | **0.65** | 0.59$^*$ (±0.02) | 0.67 | 0.54 | 0.60 | 0.57$^*$ (±0.01) |
| XLM-R$_{trans}$ | **0.74** | 0.69 | 0.42 | 0.52 (±0.01) | 0.69 | 0.58 | 0.55 | 0.56 (±0.00) | **0.72** | 0.62 | 0.60 | 0.61$^*$ (±0.02) |
| IndicBERT$_{trans}$ | 0.71 | 0.62 | 0.38 | 0.47 (±0.04) | 0.64 | 0.50 | 0.63 | 0.56$^*$ (±0.02) | 0.70 | 0.59 | 0.57 | 0.58$^*$ (±0.04) |
| MuRIL$_{trans}$ | 0.73 | 0.65 | 0.43 | 0.52 (±0.02) | 0.68 | 0.55 | 0.67 | **0.60**$^*$ (±0.01) | **0.72** | 0.60 | 0.66 | **0.63**$^*$ (±0.03) |

**Fig. 2.** Visualisation of attention scores: The upper row in each sub-figure reported the attention scores when training was done only on code-mixed samples, whereas the lower row reported the attention scores when training was done with **Train-1**. We saw similar pattern for **Train-2** as well.

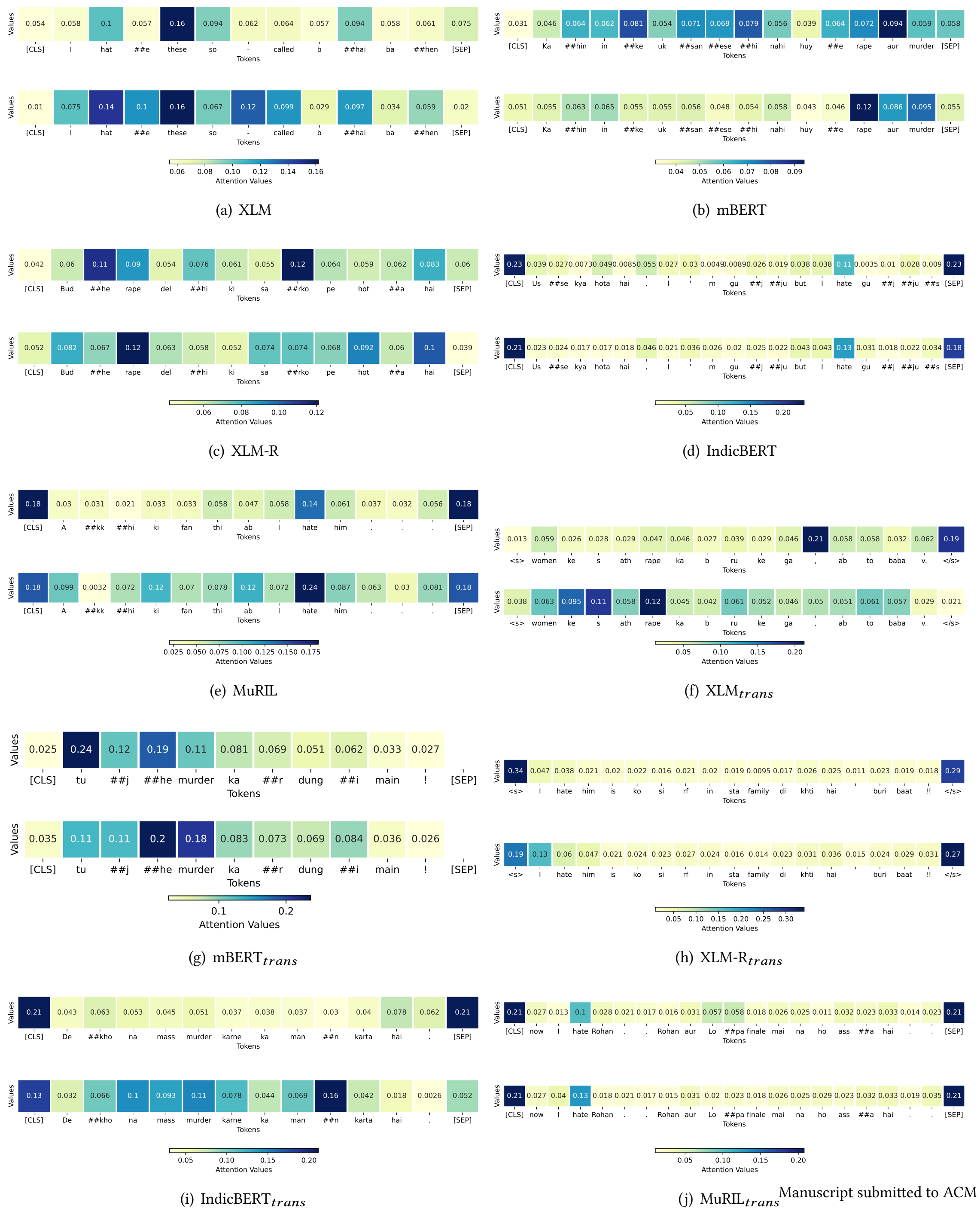

(a) XLM

(b) mBERT

(c) XLM-R

(d) IndicBERT

(e) MuRIL

(f) XLM$_{trans}$

(g) mBERT$_{trans}$

(h) XLM-R$_{trans}$

(i) IndicBERT$_{trans}$

(j) MuRIL$_{trans}$

Among MLMs, MuRIL berformed best with an $F1$-score of **0.54** followed by XLM, XLM-R$_{trans}$ and MuRIL$_{trans}$ (all with an $F1$-score of **0.52**). The performance of MLMs with and without the transformer layer remained consistent with the $F1$-score variation within **0.47** to **0.54**.

- One interesting observation is that the Naive Bayes classifier performed ($F1$ : **0.51**) nearly as good as the MLMs when we trained with only code-mixed samples. Upon inspection, we found that the lexical variation between the class samples is high for the code-mixed dataset. The KL divergence quantitatively refers to the same by reporting a high value of **1.218** for this dataset. Further, the Naive Bayes classifier is known [5] to perform better for datasets where KL divergence is high.
- On combining the native language samples, statistical models got confused and performed worse. On the contrary, MLMs significantly improved (an increase of **0.08** and **0.11** in best-reported $F1$ scores for Train-1 and Train-2, respectively). The MLMs for which the improvements were statistically significant (t-test, $p < 0.05$) are marked with an asterisk ($*$) next to their $F1$ scores. We believe this is because statistical models took n-grams as features, whereas MLMs took token embeddings for the same. The vocabulary in statistical models accommodated more Hindi and English words after we added the native/ monolingual samples to the code-mixed training set. We believe this changed the lexical distribution of n-grams as there are additional monolingual contexts. Since the statistical models primarily work with probability distributions and lexical dissimilarity among class samples, we believe their change worsened the overall performance. Past studies [55] also report similar findings.
- We observed that generic MLMs (mBERT, XLM and XLM-R) and MLMs specific to Indian languages (IndicBERT and MuRIL) reported similar F1 scores (F1-scores in the range **0.47** and **0.52**). This observation is persistent upon adding transformer layers and mixing the native samples in the training sets.
- We observed minimal or no $F1$-score improvement when transformer layers were added to the MLMs. This indicates that fine-tuning the last layers of MLMs can serve the purpose to a large extent.
- We observed no clear performance difference in MLMs when native language samples were mixed in different label ratios (**Train-1** vs **Train-2**). It indicates that they are relatively robust towards label unbalancing.
- We investigated the attention scores given by MLMs after they were trained with or without the native samples. We observed that after native sample mixing, the MLMs gave high scores to the hate-emitting words when they appeared in code-mixed contexts. Figure 2 presents some of such cases. MLMs gave more attention scores to words like 'murder' and 'rape' when they were trained with additional native samples.

### 6.2 Exp.-2 observations:

In this section, we reported the results of Exp-2, i.e. quantifying the impact of native samples when they were added in different amounts. As mentioned in section 4.4, we created several training sets by incrementally adding native sample batches (both in an equal label ratio and the label ratio same as in the original code-mixed dataset, a.k.a. code-mixed ratio). The $F1$ scores we obtained for different training sets were presented in Figure 3. The absolute values of all measured parameters were reported in Table-9. These scores are averaged over runs with three different random seeds. In the following, we reported best-performing MLMs when different batch-size native samples were added to the code-mixed training set.

- When a batch of 200 samples from both native languages were added in an equal ratio, mBERT reported the highest F1 score (**0.60**) followed by mBERT$_{trans}$ (**0.58**). When the same amount of native samples was added in a code-mixed ratio, MuRIL reported the highest F1 (**0.62**) followed by XLM, mBERT and MuRIL$_{trans}$ (**0.58**).
- On adding batches of 400 native samples in an equal ratio, mBERT again gave the highest score (**0.60**) followed by mBERT$_{trans}$ and MuRIL$_{trans}$ (with an F1 of **0.59**). When the same amount of samples was added in a code-mixed ratio, MuRIL gave the highest F1 (**0.62**) followed by mBERT (**0.59**).
- When batches of 600 native samples were added in an equal ratio, mBERT and MuRIL$_{trans}$ reported the highest F1 score (**0.59**) followed by XLM$_{trans}$ and XLM-R$_{trans}$ (**0.58**). When the same amount of samples was added in the code-mixed ratio, MuRIL performed best with an F1 score of **0.62** followed by MuRIL$_{trans}$ with an F1 score **0.61**.
- When 800 native samples were added in an equal ratio of labels, mBERT$_{trans}$ outperformed others with the highest F1 score of **0.59** followed by IndicBERT and IndicBERT$_{trans}$ (**0.58**). When the same amount of samples was mixed in code-mixed ratio, MuRIL reported the highest F1 (**0.63**) followed by MuRIL$_{trans}$ (**0.59**).
- When 1000 native samples were added in an equal ratio of labels, MuRIL and MuRIL$_{trans}$ gave the highest F1 score (**0.59**) followed by mBERT$_{trans}$ (**0.57**). When the same amount of samples was mixed in code-mixed ratio, again MuRIL and MuRIL$_{trans}$ reported the highest F1 (**0.61**) followed by IndicBERT and XLM$_{trans}$ (**0.58**).
- When 1200 native samples were added in an equal ratio of labels, IndicBERT, IndicBERT$_{trans}$ and MuRIL$_{trans}$ gave the highest F1 score (**0.60**) followed by XLM$_{trans}$ (**0.59**). When the same amount of samples were mixed in code-mixed ratio, MuRIL reported the highest F1 (**0.62**) followed by MuRIL$_{trans}$ (**0.58**).
- When 1400 native samples were added in an equal ratio of labels, mBERT$_{trans}$ gave the highest F1 score (**0.60**) followed by MuRIL$_{trans}$ (**0.59**). When the same amount of samples were mixed in code-mixed ratio, again MuRIL and MuRIL$_{trans}$ reported the highest F1 (**0.62**) followed by XLM-R$_{trans}$ (**0.57**).

Some of the interesting observations we got are,

- There was no monotonical change in the $F1$-scores when we added native samples in different amounts. Further, there is no clear variation in F1-scores (i) when samples were added with equal ratios vs CM ratios and (ii) when MLMs were trained with and without transformer layers. Interestingly, MuRIL reported upper bound F1-scores for most of the added batch versions.
- While the $F1$ score stayed between **0.53** and **0.60** for equal label ratio set-up, a relatively high fluctuation, i.e. between **0.47** and **0.63**, was observed when label ratio was in code-mixed ratio. The MLMs with additional transformer layers showed relatively higher fluctuations than those without extra transformer layers. This aligns with the idea that additional transformer layers can lead to better or worse performance.
- While most of the MLMs resulted in high fluctuation of F1 scores (upto 0.11 for mBERT$_{trans}$ with CM ratio), MuRIL consistently outperformed them with minimal variation.

### 6.3 Exp.-3 observations:

In this section, we reported the results of our experiment evaluating the performance of MLMs when they were trained with only native samples. As described in section 4.5, we trained the MLMs using (i) Hindi samples, (ii) English samples, and (iii) both English and Hindi samples together. We kept the equal label ratios in all of the training sets. This is because we didn't observe any significant $F1$-score variations in the previous experiments by not keeping the same. The $F1$-scores of MLMs for different training sets were reported in Table 10. We observed the following,

**Fig. 3.** Results of Experiments quantifying the impact of native samples when they were added in different amounts. The $x - axis$ represents the native sample batch size, and the $y - axis$ represents the $F1$ scores. The scores are averaged over three random seeds.

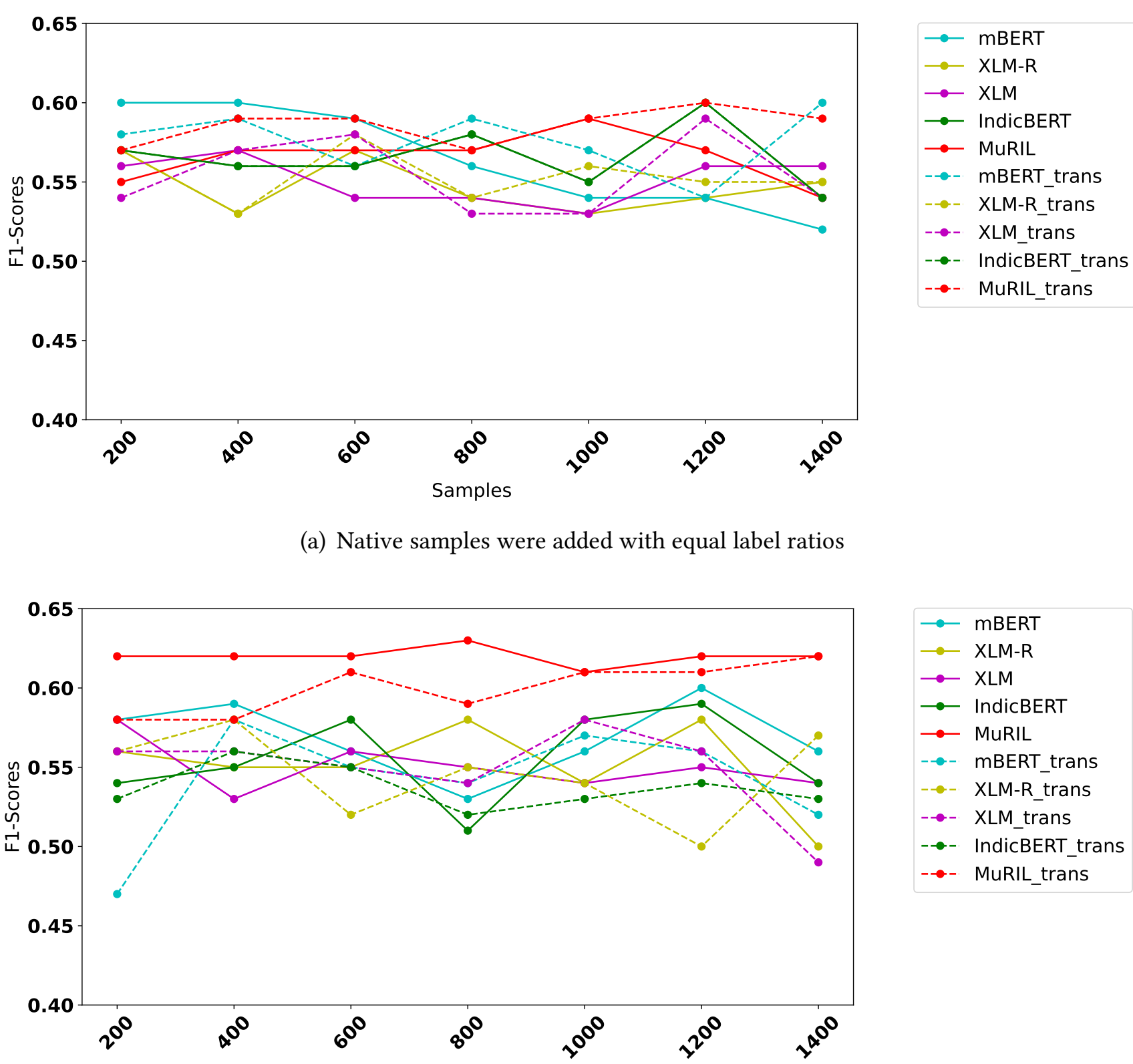

(a) Native samples were added with equal label ratios

(b) Native samples were added in a ratio as observed in the original code-mixed training set

- The highest $F1$-score we got when models were trained using only English samples was **0.54**, while the same with Hindi samples raised to **0.60**.We found this interesting because all Hindi samples were in the Devanagari script, while most code-mixed test samples were in the Roman script. We believe there could be two possible reasons. The first one is based on the observation that the code-mixed hate dataset is majorly Hindi dominant with nearly **39** percent non-dominant (majorly English) words. Therefore, when MLMs were trained with Hindi samples, they could capture the code-mixed hate. The other possible reason could be that both Hindi and code-mixed samples came from the same cultural context. Both datasets were labelled by annotators who know Hindi; hence, the models possibly captured the common cultural context. Some of the past works [6, 26] also reported similar phenomena, arguing that cross-lingual learning results by MLMs were good only when training samples and test samples came from similar languages and their cultural context were similar.
- The variants of the XLM-R language model gave the highest $F1$ scores across all native training scenarios; this means the projections of XLM-R generalise better from the native context to the code-mixed context.

**Table 9.** The experiment results quantifying the impact of native samples when they were added in different amounts. The reported values under columns 200, 400, ... up to 1400 were obtained when training sets were $\text{Train}_{200}$, $\text{Train}_{400}$, ... $\text{Train}_{1400}$ respectively. The highest parameter values across training sets were marked in bold. The highest parameter values across training sets are marked in **bold** and across models (compared separately in MLMs and $\text{MLM}_{trans}$s) are marked in blue.

| Models | Ratio | Scores | Samples size | | | | | | |
|---|---|---|---|---|---|---|---|---|---|
| | | | **200** | **400** | **600** | **800** | **1000** | **1200** | **1400** |
| XLM | Equal ratio | ACC | 0.66 | 0.67 | 0.69 | **0.70** | 0.66 | 0.67 | 0.67 |
| | | PRE | 0.53 | 0.54 | 0.57 | **0.60** | 0.53 | 0.54 | 0.54 |
| | | REC | **0.59** | **0.59** | 0.52 | 0.49 | 0.53 | 0.58 | 0.58 |
| | | F1 | 0.56 | **0.57** | 0.54 | 0.54 | 0.53 | 0.56 | 0.54 |
| | CM ratio | ACC | 0.69 | **0.70** | 0.68 | 0.68 | 0.69 | **0.70** | 0.68 |
| | | PRE | 0.57 | **0.61** | 0.56 | 0.56 | 0.58 | **0.61** | 0.56 |
| | | REC | **0.59** | 0.46 | 0.57 | 0.55 | 0.50 | 0.51 | 0.53 |
| | | F1 | **0.58** | 0.53 | 0.56 | 0.55 | 0.54 | 0.55 | 0.54 |
| mBERT | Equal ratio | ACC | 0.68 | 0.70 | **0.71** | **0.71** | 0.69 | 0.66 | 0.65 |
| | | PRE | 0.56 | 0.58 | **0.62** | **0.62** | 0.59 | 0.53 | 0.51 |
| | | REC | **0.64** | 0.63 | 0.56 | 0.50 | 0.49 | 0.55 | 0.53 |
| | | F1 | **0.60** | **0.60** | 0.59 | 0.56 | 0.54 | 0.54 | 0.56 |
| | CM ratio | ACC | 0.69 | 0.67 | 0.68 | **0.73** | 0.72 | 0.72 | 0.68 |
| | | PRE | 0.56 | 0.54 | 0.56 | **0.71** | 0.66 | 0.61 | 0.57 |
| | | REC | 0.60 | **0.64** | 0.55 | 0.42 | 0.49 | 0.60 | 0.56 |
| | | F1 | 0.58 | 0.59 | 0.56 | 0.53 | 0.56 | **0.60** | 0.56 |
| XLM-R | Equal ratio | ACC | **0.71** | 0.70 | 0.68 | 0.70 | 0.70 | 0.69 | 0.68 |
| | | PRE | 0.61 | 0.60 | 0.56 | **0.62** | **0.62** | 0.58 | 0.56 |
| | | REC | 0.54 | 0.48 | **0.58** | 0.49 | 0.46 | 0.50 | 0.53 |
| | | F1 | **0.57** | 0.53 | **0.57** | 0.54 | 0.53 | 0.54 | 0.55 |
| | CM ratio | ACC | 0.68 | **0.71** | 0.70 | 0.70 | 0.69 | 0.66 | 0.69 |
| | | PRE | 0.56 | **0.62** | 0.60 | 0.59 | 0.59 | 0.52 | 0.60 |
| | | REC | 0.57 | 0.50 | 0.50 | 0.57 | 0.49 | **0.64** | 0.43 |
| | | F1 | 0.56 | 0.55 | 0.55 | **0.58** | 0.54 | **0.58** | 0.50 |
| IndicBERT | Equal ratio | ACC | 0.67 | 0.67 | 0.67 | **0.68** | 0.64 | 0.67 | 0.61 |
| | | PRE | 0.54 | 0.54 | **0.55** | **0.55** | 0.50 | 0.54 | 0.47 |
| | | REC | 0.59 | 0.58 | 0.57 | 0.60 | 0.61 | **0.67** | 0.63 |
| | | F1 | 0.57 | 0.56 | 0.56 | 0.58 | 0.55 | **0.60** | 0.54 |
| | CM ratio | ACC | 0.65 | 0.68 | 0.68 | 0.66 | **0.72** | 0.68 | 0.64 |
| | | PRE | 0.51 | 0.56 | 0.56 | 0.53 | **0.65** | 0.56 | 0.50 |
| | | REC | 0.58 | 0.54 | 0.60 | 0.50 | 0.51 | **0.63** | 0.59 |
| | | F1 | 0.54 | 0.55 | 0.58 | 0.51 | 0.58 | **0.59** | 0.54 |
| MuRIL | Equal ratio | ACC | 0.68 | 0.64 | 0.67 | 0.67 | **0.70** | 0.69 | 0.63 |
| | | PRE | 0.56 | 0.50 | 0.54 | 0.55 | **0.58** | 0.57 | 0.49 |
| | | REC | 0.55 | **0.67** | 0.59 | 0.58 | 0.60 | 0.56 | 0.60 |
| | | F1 | 0.55 | 0.57 | 0.57 | 0.57 | **0.59** | 0.57 | 0.54 |
| | CM ratio | ACC | 0.71 | 0.72 | **0.74** | 0.72 | 0.70 | 0.70 | 0.70 |
| | | PRE | 0.58 | 0.62 | **0.65** | 0.60 | 0.58 | 0.57 | 0.58 |
| | | REC | 0.66 | 0.61 | 0.60 | **0.67** | 0.64 | **0.67** | 0.66 |
| | | F1 | 0.62 | 0.62 | 0.62 | **0.63** | 0.61 | 0.62 | 0.62 |

| Models | Ratio | Scores | Samples size | | | | | | |
|---|---|---|---|---|---|---|---|---|---|
| | | | **200** | **400** | **600** | **800** | **1000** | **1200** | **1400** |
| $\text{XLM}_{trans}$ | Equal ratio | ACC | 0.68 | **0.71** | 0.68 | 0.69 | **0.71** | 0.68 | 0.67 |
| | | PRE | 0.56 | 0.61 | 0.55 | 0.60 | **0.63** | 0.55 | 0.56 |
| | | REC | 0.53 | 0.53 | 0.62 | 0.47 | 0.46 | **0.63** | 0.52 |
| | | F1 | 0.54 | 0.57 | 0.58 | 0.53 | 0.53 | **0.59** | 0.56 |
| | CM ratio | ACC | **0.70** | **0.70** | 0.69 | 0.68 | 0.69 | **0.70** | 0.65 |
| | | PRE | 0.60 | **0.61** | 0.58 | 0.56 | 0.57 | 0.59 | 0.52 |
| | | REC | 0.53 | 0.51 | 0.52 | 0.52 | **0.59** | 0.53 | 0.46 |
| | | F1 | 0.56 | 0.56 | 0.55 | 0.54 | **0.58** | 0.56 | 0.49 |
| $\text{mBERT}_{trans}$ | Equal ratio | ACC | 0.69 | **0.70** | 0.66 | 0.69 | 0.66 | 0.68 | 0.69 |
| | | PRE | 0.56 | **0.59** | 0.52 | 0.57 | 0.52 | 0.57 | 0.56 |
| | | REC | 0.61 | 0.59 | 0.59 | 0.61 | 0.63 | 0.51 | **0.64** |
| | | F1 | 0.58 | 0.59 | 0.56 | 0.59 | 0.57 | 0.54 | **0.60** |
| | CM ratio | ACC | **0.71** | 0.69 | 0.69 | **0.71** | 0.69 | 0.65 | 0.70 |
| | | PRE | **0.68** | 0.56 | 0.58 | 0.65 | 0.58 | 0.52 | 0.63 |
| | | REC | 0.36 | 0.60 | 0.51 | 0.47 | 0.55 | **0.61** | 0.45 |
| | | F1 | 0.47 | **0.58** | 0.55 | 0.54 | 0.57 | 0.56 | 0.52 |
| $\text{XLM-R}_{trans}$ | Equal ratio | ACC | 0.68 | **0.70** | 0.69 | **0.70** | **0.70** | **0.70** | 0.69 |
| | | PRE | 0.55 | **0.61** | 0.56 | **0.61** | 0.60 | 0.60 | 0.58 |
| | | REC | 0.59 | 0.47 | **0.61** | 0.48 | 0.52 | 0.51 | 0.52 |
| | | F1 | **0.57** | 0.53 | 0.58 | 0.54 | 0.56 | 0.55 | 0.55 |
| | CM ratio | ACC | 0.69 | 0.69 | **0.70** | **0.70** | 0.67 | **0.70** | 0.69 |
| | | PRE | 0.58 | 0.57 | 0.63 | 0.61 | 0.54 | **0.65** | 0.57 |
| | | REC | 0.54 | **0.60** | 0.44 | 0.51 | 0.54 | 0.40 | 0.57 |
| | | F1 | 0.56 | **0.58** | 0.52 | 0.55 | 0.54 | 0.50 | 0.57 |
| $\text{IndicBERT}_{trans}$ | Equal ratio | ACC | 0.67 | 0.67 | 0.67 | **0.68** | 0.64 | 0.67 | 0.61 |
| | | PRE | 0.54 | 0.54 | **0.55** | **0.55** | 0.50 | 0.54 | 0.47 |
| | | REC | 0.59 | 0.58 | 0.57 | 0.60 | 0.61 | **0.67** | 0.63 |
| | | F1 | 0.57 | 0.56 | 0.56 | 0.58 | 0.55 | **0.60** | 0.54 |
| | CM ratio | ACC | 0.67 | **0.72** | 0.69 | 0.63 | 0.67 | 0.71 | 0.71 |
| | | PRE | 0.56 | **0.64** | 0.59 | 0.49 | 0.56 | **0.64** | 0.63 |
| | | REC | 0.50 | 0.49 | 0.51 | **0.54** | 0.51 | 0.47 | 0.45 |
| | | F1 | 0.53 | **0.56** | 0.55 | 0.52 | 0.53 | 0.54 | 0.53 |
| $\text{MuRIL}_{trans}$ | Equal ratio | ACC | 0.67 | 0.67 | 0.70 | **0.71** | 0.66 | 0.68 | 0.67 |
| | | PRE | 0.54 | 0.53 | 0.58 | **0.61** | 0.53 | 0.55 | 0.53 |
| | | REC | 0.59 | 0.66 | 0.61 | 0.53 | **0.67** | **0.67** | **0.67** |
| | | F1 | 0.57 | 0.59 | 0.59 | 0.57 | 0.59 | **0.60** | 0.59 |
| | CM ratio | ACC | 0.70 | 0.70 | 0.69 | 0.69 | 0.70 | 0.68 | **0.71** |
| | | PRE | **0.59** | **0.59** | 0.57 | 0.57 | **0.59** | 0.55 | **0.59** |
| | | REC | 0.57 | 0.57 | 0.66 | 0.62 | 0.63 | **0.67** | 0.65 |
| | | F1 | 0.58 | 0.58 | 0.61 | 0.59 | 0.61 | 0.61 | **0.62** |

**Table 10.** $F1$-scores of MLMs when they were trained with only native samples. The variations in $F1$ scores for different random seeds were reported in brackets.

| Models | Training sets | | |
|---|---|---|---|
| | **English** | **Hindi** | **English + Hindi** |
| **XLM** | 0.47 (±0.03) | 0.34 (±0.04) | 0.36 (±0.02) |
| **mBERT** | 0.50 (±0.04) | 0.41 (±0.03) | 0.51 (±0.01) |
| **XLM-R** | 0.53 (±0.01) | **0.60** (±0.03) | **0.60** (±0.01) |
| **IndicBERT** | 0.50 (±0.02) | 0.41 (±0.05) | 0.52 (±0.00) |
| **MuRIL** | 0.53 (±0.01) | 0.13 (±0.02) | 0.46 (±0.07) |
| $\textbf{XLM}_{trans}$ | 0.36 (±0.04) | 0.21 (±0.05) | 0.45 (±0.04) |
| $\textbf{mBERT}_{trans}$ | 0.48 (±0.02) | 0.31 (±0.04) | 0.51 (±0.02) |
| $\textbf{XLM-R}_{trans}$ | **0.54** (±0.00) | 0.55 (±0.02) | 0.57 (±0.03) |
| $\textbf{IndicBERT}_{trans}$ | 0.49 (±0.03) | 0.45 (±0.06) | 0.53 (±0.01) |
| $\textbf{MuRIL}_{trans}$ | **0.54** (±0.01) | 0.34 (±0.05) | 0.47 (±0.04) |

- There were no significant improvements in the best $F1$ scores when Hindi and English samples were trained together, compared to when training was done with only Hindi samples. Also, the highest $F1$ score we got by training on **Train-1** and **Train-2** (of Exp-1) sets was **0.63**, which is a minimal improvement over when training was done with native samples (**0.60**). This empirically points out that MLMs trained on native samples can predict code-mixed hate to a large extent.

## 7 ERROR ANALYSIS:

In this section, we reported the cases in which combining the native samples improved or degraded the performance of considered MLMs. Table 11 reported some of such samples. We observed the following:

- The addition of the native samples helped the MLMs to identify hate, expressed in a code-mixed phrase, without using any explicit hate words. Examples reported in *Sl. No.* 1 and 2 demonstrate this. Even though there is no explicit hate word mentioned in *'tum logo ne hi karwaya tha blast..'* (**Gloss:** You guys have done blast..), most MLMs trained with added native samples could detect the hate. Similarly, in *SL. No. 2*, MLMs trained without native samples failed to understand that the Hindi hate word *'nafrat'* is used to convey a non-hate message.
- For the cases where hate is expressed in a sarcastic tone, we observed that MLMs generally struggle to identify it. For example, in *Sl. No.* 3, only the $XLM-R_t$ model trained **T1** and $MuRIL_t$ could identify it.
- Finally, in many cases, like *Sl. No.* 4 and 5, we saw performance degradation after adding the native samples. After a careful inspection by the expert linguists, we found that they are some of the hard cases to identify, even for the experts. This is because the hate expressed here is subjective. For instance, if we see the translation of *Sl. No.* 5 i.e. *"Brother... Indians are forgetful... they forget everything... demonetization... what happened?"*, the label assigned to it was 'Hate'; however, many linguists preferred to categorise it as a 'criticism' without having a solid hate component (E.g. if the writer is an Indian). Similarly, the sample in row 6 translates to *'there is still time, improve yourself'* can be considered a case of implicit hate (patronizing and condescending language) depending on the context.

## 8 CONCLUSIONS:

In this paper, we presented several experiments to evaluate the impact of native language samples on code-mixed hate detection. Some of the important observations we got were,

- On combining the native hate samples in the code-mixed training set, we found that MLMs performed relatively better (an increase of **0.11** in the F1 score for the best-performing model i.e. $MuRIL_{trans}$). For many of them, the improvements were statistically significant. The attention scores produced by MLMs also validated this fact.
- MLMs trained with only native samples could identify hate in the code-mixed context to a large extent (**0.60** in only native samples vs **0.63** in combined). It implies that we can deploy them if an appropriate code-mixed corpus is unavailable. We also observed that the native corpus (here Hindi) that shares maximum linguistic overlap with the code-mixed hate corpus is more effective in capturing code-mixed hate (F1 of **0.54** for English vs F1 of **0.60** in Hindi).

## 9 FUTURE WORKS:

In this section, we report the limitations of the present work. They can be addressed in subsequent future works.

- In this work, we have evaluated our hypothesis only for a single code-mixed environment (Hindi-English) and a task (hate). This is because we don't have linguistic expertise for other code-mixed environments for which we have code-mixed hate corpora publicly available (e.g. Sinhala-English hate dataset [4]). In future, one can evaluate the idea of native sample mixing for other code-mixed environments and tasks (humour, sarcasm, etc.) for discovering its wide usage.

**Table 11.** Selected examples for various cases reported under error analysis. Here, the check-mark, and the cross-mark denote correct and incorrect classification by the corresponding model, respectively. Notation: T1 for Train-1 and T2 for Train-2. The values next to **CM** refer to the results when the models were trained with code-mixed samples.

| Sl No | Sample | Train | Models | | | | | | | | | |
|---|---|---|---|---|---|---|---|---|---|---|---|---|
| | | | **XLM** | **XLM$_t$** | **mBERT** | **mBERT$_t$** | **XLM-R** | **XLM-R$_t$** | **IndicBERT** | **IndicBERT$_t$** | **MuRIL** | **MuRIL$_t$** |
| 1 | Tum logon ne hi karwaya tha blasts in PAKISTAN iss liye aise posts daal rahe ho (**Hate**)<br>**Translation:** You guys were the ones who orchestrated the blasts in Pakistan, that's why you're posting such things. | CM | ✗ | ✗ | ✗ | ✗ | ✗ | ✗ | ✗ | ✗ | ✗ | ✓ |
| | | T1 | ✗ | ✓ | ✓ | ✓ | ✓ | ✓ | ✓ | ✗ | ✓ | ✓ |
| | | T2 | ✓ | ✗ | ✓ | ✓ | ✓ | ✗ | ✓ | ✓ | ✓ | ✓ |
| 2 | Kabhi nafrat to kabhi dilo ka mail hai, (**Non-hate**)<br>**Translation:** Sometimes it's hatred, sometimes it's the connection of hearts, | CM | ✗ | ✗ | ✗ | ✗ | ✗ | ✗ | ✗ | ✗ | ✗ | ✗ |
| | | T1 | ✓ | ✓ | ✓ | ✓ | ✓ | ✓ | ✓ | ✗ | ✓ | ✓ |
| | | T2 | ✓ | ✓ | ✓ | ✓ | ✓ | ✓ | ✗ | ✓ | ✗ | ✓ |
| 3 | Tabhi Loktantra ka Murder BJP karti hai Supreme court bhi kai bar Phatkaar laga Chuka hai (**Hate**)<br>**Translation:** That's why BJP commits the murder of democracy; the Supreme Court has also reprimanded them many times. | CM | ✗ | ✗ | ✗ | ✗ | ✗ | ✗ | ✗ | ✗ | ✗ | ✗ |
| | | T1 | ✗ | ✗ | ✗ | ✗ | ✗ | ✓ | ✗ | ✗ | ✓ | ✓ |
| | | T2 | ✗ | ✗ | ✗ | ✗ | ✗ | ✗ | ✗ | ✗ | ✗ | ✓ |
| 4 | Bhai...Indian bhulakkar hote he ...San bhul hate he...notebandhhi...kya hua? (**Hate**)<br>**Translation:** Brother... Indians are forgetful... They forget everything... Demonetization... what happened? | CM | ✓ | ✓ | ✗ | ✓ | ✗ | ✓ | ✓ | ✓ | ✗ | ✗ |
| | | T1 | ✗ | ✗ | ✗ | ✗ | ✗ | ✗ | ✗ | ✗ | ✗ | ✗ |
| | | T2 | ✗ | ✓ | ✓ | ✗ | ✗ | ✓ | ✗ | ✗ | ✗ | ✗ |
| 5 | Abhi bhe time he sab sudar jao. (**Non-hate**)<br>**Translation:** There's still time. Everyone, mend your ways. | CM | ✗ | ✓ | ✗ | ✗ | ✓ | ✓ | ✓ | ✓ | ✓ | ✗ |
| | | T1 | ✗ | ✗ | ✓ | ✗ | ✗ | ✗ | ✓ | ✗ | ✗ | ✓ |
| | | T2 | ✗ | ✓ | ✗ | ✓ | ✓ | ✓ | ✗ | ✓ | ✗ | ✗ |

- We have kept the architecture and the task setup simple and basic. We believe a wider testing of our hypothesis across many task set-ups (e.g. multi-tasking, sequence labelling, sequence-to-sequence tasks, etc.) can be done to report further the applicability of the idea of native language mixing. For example, in this paper, we found that for the cases where code-mixed hate was sarcastic or subjective, it's still hard to identify them by simply including native hate samples. One can check in future if multi-tasking with native and code-mixed sarcasm samples helps the models to perform better.

[4] https://huggingface.co/datasets/NLPC-UOM/Sinhala-English-Code-Mixed-Code-Switched-Dataset

- We could show our results only for a few language models, i.e. XLM, mBERT, XLM-R, IndicBERT and MuRIL. Expanding the experiments to newer language models will showcase the strength and relevance of the idea of native language mixing in more recent MLMs.
- Hate in one language may be "banter between friends" in code-mixed situations. In the current study, we have not incorporated any mechanisms to handle such cases. However, such aspects can be studied in the future. Further, in our understanding, there are no publicly available datasets that have annotations of hate intensity levels when utterances are made in code-mixed and monolingual language. Creating such a dataset in future can be useful to the NLP community.